%% file: sigconf_main.tex
\documentclass[sigconf]{acmart}
\usepackage{utils}

\copyrightyear{2022}
\acmYear{2022}
\setcopyright{acmlicensed}\acmConference[GECCO '22]{Genetic and
Evolutionary Computation Conference}{July 9--13, 2022}{Boston, MA, USA}
\acmBooktitle{Genetic and Evolutionary Computation Conference (GECCO '22),
July 9--13, 2022, Boston, MA, USA}
\acmPrice{15.00}
\acmDOI{10.1145/3512290.3528827}
\acmISBN{978-1-4503-9237-2/22/07}

\begin{document}

\title{CMA-ES with Margin: Lower-Bounding Marginal Probability for Mixed-Integer Black-Box Optimization}


\author{Ryoki Hamano}
\affiliation{%
  \institution{Yokohama National University}
  \city{Yokohama}
  \state{Kanagawa}
  \country{Japan}
}
\email{hamano-ryoki-pd@ynu.jp}

\author{Shota Saito}
\affiliation{%
  \institution{Yokohama National University \& SkillUp AI Co., Ltd.}
  \city{Yokohama}
  \state{Kanagawa}
  \country{Japan}
}
\email{saito-shota-bt@ynu.jp}

\author{Masahiro Nomura}
\affiliation{%
  \institution{CyberAgent}
  \city{Shibuya}
  \state{Tokyo}
  \country{Japan}
 }
\email{nomura\_masahiro@cyberagent.co.jp}

\author{Shinichi Shirakawa}
\affiliation{%
  \institution{Yokohama National University}
  \city{Yokohama}
  \state{Kanagawa}
  \country{Japan}
}
\email{shirakawa-shinichi-bg@ynu.ac.jp}


    \renewcommand{\shortauthors}{Hamano and Saito et al.}

\begin{abstract}
This study targets the mixed-integer black-box optimization (MI-BBO) problem where continuous and integer variables should be optimized simultaneously. The CMA-ES, our focus in this study, is a population-based stochastic search method that samples solution candidates from a multivariate Gaussian distribution (MGD), which shows excellent performance in continuous BBO. The parameters of MGD, mean and (co)variance, are updated based on the evaluation value of candidate solutions in the CMA-ES. If the CMA-ES is applied to the MI-BBO with straightforward discretization, however, the variance corresponding to the integer variables becomes much smaller than the granularity of the discretization before reaching the optimal solution, which leads to the stagnation of the optimization. In particular, when binary variables are included in the problem, this stagnation more likely occurs because the granularity of the discretization becomes wider, and the existing modification to the CMA-ES does not address this stagnation. To overcome these limitations, we propose a simple modification of the CMA-ES based on lower-bounding the marginal probabilities associated with the generation of integer variables in the MGD. The numerical experiments on the MI-BBO benchmark problems demonstrate the efficiency and robustness of the proposed method. 
\end{abstract}



\maketitle
\section{Introduction}
The mixed-integer black-box optimization (MI-BBO) problem is the problem of simultaneously optimizing continuous and integer variables under the condition that the objective function is not differentiable and not available in the explicit functional form.
The MI-BBO problems often appear in real-world applications such as, material design~\cite{zhang_bayesian_2020, iyer_data_2020}, topology optimization~\cite{yang_evolutionary_1998, fujii_cma-es-based_2018}, placement optimization for $\mathrm{CO_{2}}$ capture and storage~\cite{Miyagi_GECCO2018}, and hyper-parameter optimization of machine learning~\cite{hutter_automated_2019,hazan2018hyperparameter}.
Several algorithms have been designed for MI-BBO so far, e.g., the extended evolution strategies~\cite{li_mixed_2013} and surrogate model-based method~\cite{Bliek_GECCO21}. However, despite the high demand for an efficient MI-BBO method, the BBO methods for mixed-integer problems are not actively developed compared to those for continuous or discrete problems.
One common way is applying a continuous BBO method to an MI-BBO problem by discretizing the continuous variables when evaluating a candidate solution rather than using a specialized method for MI-BBO.

The \textit{covariance matrix adaptation evolution strategy} (CMA-ES) \cite{hansen_adapting_1996, hansen_reducing_2003} is a powerful method in continuous black-box optimization, aiming to minimize the objective function through population-based stochastic search. The CMA-ES samples several continuous vectors from a multivariate Gaussian distribution (MGD) and then evaluates the objective function values of the vectors. Subsequently, the CMA-ES facilitates optimization by updating the mean vector, covariance matrix, and step-size (overall standard deviation) based on evaluation values. The CMA-ES exhibits two attractive properties for users. First, it has several invariance properties, such as the invariance of a strictly monotonic transformation of the objective function and an affine transformation (rotation and translation) of the search space. These invariances make it possible to generalize the powerful empirical performance of the CMA-ES of one particular problem to another. Second, the CMA-ES is a quasi-parameter-free algorithm, which allows use without tuning the hyperparameters, such as the learning rate for the mean vector or covariance matrix. In the CMA-ES, all hyperparameters are given default values based on theoretical work and careful experiments.

The most straightforward way to apply the CMA-ES to the MI-BBO is to discretize some elements of the sampled continuous vector, e.g., \cite{TAMILSELVI2014208}. However, because of the plateau caused by the discretization, this simple method may not change the evaluation value by small variations in elements corresponding to the integer variables. Specifically, as pointed out in \cite{hansen_cma-es_2011}, this stagnation occurs when the sample standard deviation of the dimension corresponding to an integer variable becomes much smaller than the granularity of the discretization. In particular, the step-size tends to decrease with each iteration, which promotes trapping on the plateau of the integer variables.
To address this plateau problem in the integer variable treatment, \citet{hansen_cma-es_2011} proposed the injection of mutations into a sample of elements corresponding to integer variables in the CMA-ES, and \citet{Miyagi_GECCO2018} used this modification for the real-world MI-BBO problem.
Although this mutation injection is effective on certain problem classes, \citet{hansen_cma-es_2011} mentioned that it is not suitable for binary variables or $k$-ary integers in $k < 10$.

This study aims to improve the integer variable treatment of the CMA-ES in the MI-BBO problem. First, we investigate why the CMA-ES search fails in MI-BBO problems involving binary variables. Following the result, we propose the adaptation of the sample discretization process according to the current MGD parameters. The proposed adaptive discretization process can be represented as an affine transformation for a sample. Therefore, owing to the affine invariance of CMA-ES, it is expected to maintain the good behavior of the original CMA-ES. Additionally, we extend the proposed method from binary variables to integer variables.

\begin{table}[t]
    \centering
    \caption{Default hyperparameters and initial values of the CMA-ES.} \vspace{-2.5mm}
    \begin{tabular}{c|c}
        \hline
        $\lambda$ & 4 + $\lfloor 3 \ln(N) \rfloor$ \\
        $\mu$ & $\lfloor \lambda / 2 \rfloor$ \\
        \hline
        $w'_{i}$ & $ \ln\left( \frac{\lambda + 1}{2} \right) - \ln i$ \\
        $w_{i} (i \leq \mu)$ & $w'_{i} \left( \sum_{j=1}^\mu w'_{i} \right)^{-1}$ \\
        $w_{i} (i > \mu)$ & $\frac{ w'_{i} }{\sum_{j=\mu + 1}^\lambda | w'_{i} | } \min \left( 1 + \frac{c_1}{c_\mu}, 1 + \frac{2 \muw^{-}}{\muw + 2}, \frac{1 - c_1 - c_\mu}{N c_\mu} \right)$  \\
        \hline
        $\muw$ & $\left( \sum_{j=1}^\mu (w'_{i})^2 \right)^{-1}$ \\
        $\muw^{-}$ & $\left(\sum_{j=\mu + 1}^\lambda w'_{i} \right)^2  \left(\sum_{j=\mu + 1}^\lambda (w'_{i})^2 \right)^{-1}$ \\
        \hline
        $c_m$ & 1 \\
        $c_{\sigma}$ & $(\muw + 2) / (N + \muw + 5)$ \\ 
        $c_{c}$ & $(4 + \muw / N) / (N + 4 + 2\muw / N)$ \\ 
        $c_{1}$ & $2\left((N + 1.3)^2 + \muw \right)^{-1}$ \\ 
        $c_{\mu}$ & $\min \left( 1 - c_1 , \frac{2(\muw - 2 + 1 / \muw)}{(N + 2)^2 + \muw} \right)$ \\
        $d_{\sigma}$ & $1 + c_{\sigma}+ 2 \max\left(0, \sqrt{\frac{\muw - 1}{N + 1}} - 1 \right)$ \\
        \hline
        $\ps[0]$, $\pc[0]$ & $\boldsymbol{0}$ \\
        $\m[0], \C[0], \sig[0]$ & Depending on the problem \\
        \hline
    \end{tabular}
    \label{tb:cma_params}
\end{table}
\vspace{-1mm}
\section{CMA-ES and Mixed Integer Handling}
\label{sec:cma-es}
\subsection{CMA-ES}
Let us consider the black-box minimization problem in the continuous search space for an objective function $f: \R^{N} \rightarrow \R$. The CMA-ES samples an $N$-dimensional candidate solution $\x \in \R^{N}$ from an MGD $\mathcal{N}(\boldsymbol{m}, \sigma^2 \boldsymbol{C})$ parameterized by the mean vector $\boldsymbol{m} \in \R^{N}$, covariance matrix $\boldsymbol{C} \in \R^{N \times N}$, and step-size $\sigma \in \R_{> 0}$. The CMA-ES updates the distribution parameters based on the objective function value $f(\x)$. There are several variations in the update methods of distribution parameters, although we consider the de facto standard CMA-ES \cite{hansen2016cma}, which combines the weighted recombination, cumulative step-size adaptation, rank-one covariance matrix update, and rank-$\mu$ update. We use the default parameters proposed in \cite{hansen2016cma} and listed them in Table \ref{tb:cma_params}. The CMA-ES repeats the following steps until a termination criterion is satisfied.

\paragraph{Sample and Evaluate Candidate Solutions}
In the $t$-th iteration,  the $\lambda$ candidate solutions $\boldsymbol{x}_{i}$ ($i = 1, 2, \dots, \lambda$) are sampled independently from the MGD $\mathcal{N}(\m[t], (\sig[t])^{2} \C[t])$ as follows:
\begin{align}
    \y_i = (\C[t])^{\frac{1}{2}} \boldsymbol{\xi}_{i} \enspace, \\
    \x_i = \m[t] + \sig[t] \y_i \enspace,
\end{align}
where $\boldsymbol{\xi}_i \sim \mathcal{N}(\boldsymbol{0}, \boldsymbol{I})$ represents a random vector with zero mean and a covariance matrix of the identity matrix $\boldsymbol{I}$, and $(\C[t])^{\frac{1}{2}}$ is the square root of the covariance matrix $\C[t]$ that is the symmetric and positive definite matrix satisfying $\C[t] = (\C[t])^{\frac{1}{2}} (\C[t])^{\frac{1}{2}}$.
The candidate solutions $\{\x_1, \x_2 \dots, x_{\lambda}\}$ are evaluated by $f$ and sorted by ranking. Let $x_{i:\lambda}$ be the $i$-th best candidate solution; then, $f(\x_{1:\lambda}) \leq f(\x_{2:\lambda}) \leq \dots \leq f(\x_{\lambda:\lambda})$ and let $\y_{i:\lambda}$ be the random vector corresponding to $\x_{i:\lambda}$. 

\paragraph{Update Mean Vector} The mean vector update uses the weighted sum of the best $\mu < \lambda$ candidate solutions and updates $\m[t]$ as follows:
\begin{align}
    \m[t+1] = \m[t] + c_{m} \sum_{i=1}^{\mu} w_{i} ( \x_{i:\lambda} - \m[t] ) \enspace, \label{eq:cma-es-m}
\end{align}
where $c_{m}$ is the learning rate for the mean vector, and the weight $w_{i}$ satisfies $w_{1} \geq w_{2} \geq \dots \geq w_{\mu} > 0$ and $\sum_{i=1}^{\mu} w_{i} = 1$.

\paragraph{Compute Evolution Paths}
For the step-size adaptation and the rank-one update of the covariance matrix, we use evolution paths that accumulate an exponentially fading pathway of the mean vector in the generation sequence. Let  $\boldsymbol{p}_{\sigma}$ and $\boldsymbol{p}_{c}$ describe the evolution paths for the step-size adaptation and rank-one update, respectively; then, $\ps[t]$ and $\pc[t]$ are updated as follows:
\begin{align}
    \ps[t+1] &= (1-c_\sigma)\ps[t] + \sqrt{c_\sigma(2-c_\sigma)\muw} {\C[t]}^{-\frac12} \sum_{i=1}^\mu w_i \y_{i:\lambda}  \enspace, \\
    \pc[t+1] &= (1-c_c) \pc[t] + h_\sigma \sqrt{c_c(2-c_c)\muw} \sum_{i=1}^\mu w_i \y_{i:\lambda} \enspace,
\end{align}
where $c_\sigma$ and $c_c$ are cumulative rates, and
{\small
\begin{align*}
    h_\sigma = \mathds{1} \left\{ \|\ps[t+1]\| < \sqrt{1-(1-c_\sigma)^{2(t+1)}}\left(1.4+\frac{2}{N+1}\right)\E [\|\mathcal{N}(\boldsymbol{0}, \boldsymbol{I}) \|] \right\}
\end{align*}
}is an indicator function used to suppress a rapid increase in $\boldsymbol{p}_{c}$, where $\E [\|\mathcal{N}(\boldsymbol{0}, \boldsymbol{I}) \|] \approx \sqrt{N} \left(1 - \frac{1}{4N} + \frac{1}{21N^2}\right)$ is the expected Euclidean norm of the sample from a standard Gaussian distribution.

\paragraph{Update Step-size and Covariance Matrix}
Using the evolution paths computed in the previous step, we update $\C[t]$ and $\sig[t]$ as follows:
\begin{align}
        \C[t+1] &= \left(1 - c_1 - c_\mu \sum^{\lambda}_{i=1} w_{i} + (1-h_\sigma)c_1 c_c(2-c_c) \right) \C[t] \notag\\ 
        &\enspace + \underbrace{c_1 \pc[t+1]{\pc[t+1]}^\top}_{\text{rank-one update}} + \underbrace{c_\mu \sum_{i=1}^\lambda w_i^{\circ} \y_{i:\lambda}\y_{i:\lambda}^\top}_{\text{rank-}\mu\text{ update}}  \enspace, \\
        \sig[t+1] &= \sig[t] \exp \left( \frac{c_\sigma}{d_\sigma} \left( \frac{\|\ps[t+1] \|}{\E [\|\mathcal{N}(\boldsymbol{0}, \boldsymbol{I}) \|]} - 1 \right) \right)  \enspace,
        \label{eq:cma-es-sig}
\end{align}
where
$w_i^{\circ} := w_i \cdot (1$ if $w_i\ge0$ else $N / \left\| (\C[t])^{-\frac{1}{2}} \boldsymbol{y}_{i:\lambda} \right\|^{2} )$, 
$c_1$ and $c_{\mu}$ are the learning rates for the rank-one and rank-$\mu$ updates, respectively. Additionally, $d_{\sigma}$ is a damping parameter for the step-size adaptation.

\subsection{CMA-ES with Mixed-Integer Handling}
\label{sec::mi-cma-es}
In \cite{hansen_cma-es_2011}, several steps of the CMA-ES are modified to handle the integer variables. To explain this modification, we apply notations $\lbrack \cdot \rbrack_{j}$ and $\langle \cdot \rangle_{j}$, where the former denotes the $j$-th element of an argument vector and the latter denotes the $j$-th diagonal element of an argument matrix. We denote the number of dimensions as $N = N_{\mathrm{co}} + N_{\mathrm{in}}$, where $N_{\mathrm{co}}$ and $N_{\mathrm{in}}$ are the numbers of the continuous and integer variables, respectively. More specifically, the 1st to $N_{\mathrm{co}}$-th elements and $(N_{\mathrm{co}} + 1)$-th to $N$-th elements of the candidate solution are the elements corresponding to the continuous and integer variables, respectively.

\paragraph{Inject Integer Mutation}
For the element corresponding to the integer variable, stagnation occurs when the sample standard deviation becomes much smaller than the granularity of the discretization. The main idea to solve this stagnation in \cite{hansen_cma-es_2011} is to inject the \emph{integer mutation vector} $\boldsymbol{r}^{\mathrm{int}}_{i} \in \mathbb{N}^{N}$ into the candidate solution, which is given by
\begin{align}
    \x_i = \m[t] + \sig[t] \y_i + \boldsymbol{S}^{\mathrm{int}} \boldsymbol{r}^{\mathrm{int}}_{i}  \enspace, 
    \label{eq::integer_mutation}
\end{align}
where $\boldsymbol{S}^{\mathrm{int}}$ is the diagonal matrix whose diagonal elements indicate the variable granularities, which is $\langle \boldsymbol{S}^{\mathrm{int}} \rangle_{j} = 1$ if $N_{\mathrm{co}} + 1 \leq j \leq N$; otherwise $\langle \boldsymbol{S}^{\mathrm{int}} \rangle_{j} = 0$ in usual case. The integer mutation vector $\boldsymbol{r}^{\mathrm{int}}_{i}$ is sampled as follows:
\begin{enumerate}[\textrm{Step }1\textrm{. }]
    \item Set up a randomly ordered set of elements indices $J^{(t)}$ satisfying $2 \sig[t] \langle \C[t] \rangle_{j}^{\frac{1}{2}} < \langle \boldsymbol{S}^{\mathrm{int}} \rangle_{j}$.
    \item Determine the number of candidate solutions into which the integer mutation is injected as follows:
    \begin{align*}
    \lambda_{\mathrm{int}}^{(t)} = \left\{
        \begin{array}{lll}
            0 & (|J^{(t)}| = 0) \\
            \min(\lambda / 10 + |J^{(t)}| + 1, \lfloor \lambda / 2 \rfloor - 1) & (0 < |J^{(t)}| < N) \\
            \lfloor \lambda / 2 \rfloor & (|J^{(t)}| = N)
        \end{array}
        \right.  \enspace .
    \end{align*}
    \item $\lbrack \boldsymbol{R}'_{i} \rbrack_{j} = 1$ if the element indicate $j$ is equal to mod$(i - 1,|J^{(t)}|)$-th element of $J^{(t)}$, otherwise $\lbrack \boldsymbol{R}'_{i} \rbrack_{j} = 0$.
    \item $\lbrack \boldsymbol{R}''_{i} \rbrack_{j}$ is sampled from a geometric distribution with the probability parameter $p = 0.7^{\frac{1}{|J^{(t)}|}}$ if $j \in J^{(t)}$, otherwise $\lbrack \boldsymbol{R}''_{i} \rbrack_{j} = 0$.
    \item $\boldsymbol{r}^{\mathrm{int}}_{i} = \pm ( \boldsymbol{R}'_{i} + \boldsymbol{R}''_{i})$ with the sign-switching probability 1/2 if $i \leq \lambda_{\mathrm{int}}^{(t)}$, otherwise $\boldsymbol{r}^{\mathrm{int}}_{i} = \boldsymbol{0}$.
    \item If $\lambda_{\mathrm{int}}^{(t)} > 0$, $\lbrack  \boldsymbol{r}^{\mathrm{int}}_{\lambda} \rbrack_{j} = \pm \left( \left\lfloor \frac{\lbrack \boldsymbol{x}_{1:\lambda}^{(t-1)} \rbrack_{j}}{\langle \boldsymbol{S}^{\mathrm{int}} \rangle_{j}} \right\rfloor - \left\lfloor \frac{\lbrack \m[t] \rbrack_{j}}{\langle \boldsymbol{S}^{\mathrm{int}} \rangle_{j}} \right\rfloor \right)$ with the sign-switching probability 1/2 if $\langle \boldsymbol{S}^{\mathrm{int}} \rangle_{j} > 0$, otherwise $\lbrack \boldsymbol{r}^{\mathrm{int}}_{\lambda} \rbrack_{j} = 0$. This is a modified version of \cite{hansen_cma-es_2011} and introduced in~\cite{Miyagi_GECCO2018}.
\end{enumerate}

\paragraph{Modify Step-size Adaptation}
If the standard deviation of the elements corresponding to the integer variables is much smaller than the granularity of the discretization, then the step-size adaptation rapidly decreases the step-size. To address this problem, \cite{hansen_cma-es_2011} proposed a modification of the step-size adaptation to remove the elements corresponding to integer variables with considerably smaller standard deviations from the evolution path $\ps[t+1]$ when updating the step-size as follows: 
\begin{align}
    \sig[t+1] &= \sig[t] \exp \left( \frac{c_\sigma}{d_\sigma} \left( \frac{\| \boldsymbol{I}_{\sigma}^{(t+1)} \ps[t+1] \|}{\E [\|\mathcal{N}(\boldsymbol{0}, \boldsymbol{I}_{\sigma}^{(t+1)} ) \|]} - 1 \right) \right)  \enspace,
\end{align}
where $\boldsymbol{I}_{\sigma}^{(t+1)}$ is the diagonal masking matrix, and $\langle \boldsymbol{I}_{\sigma}^{(t+1)} \rangle_{j} = 0$ if $5 \sigma \langle \C[t] \rangle_{j}^{\frac{1}{2}} / \sqrt{c_{\sigma}} < \langle \boldsymbol{S}^{\mathrm{int}} \rangle_{j}$; otherwise, $\langle \boldsymbol{I}_{\sigma}^{(t+1)} \rangle_{j} = 1$. The expected value $\|\mathcal{N}(\boldsymbol{0}, \boldsymbol{I}_{\sigma}^{(t+1)} ) \|$ is approximated by $\sqrt{M} \left(1 - \frac{1}{4M} + \frac{1}{21M^2}\right)$, \linebreak where $M$ is the number of non-zero diagonal elements for $\boldsymbol{I}_{\sigma}^{(t+1)}$.

\begin{figure*}[thbp]
    \centering
    \includegraphics[width=0.85\linewidth]{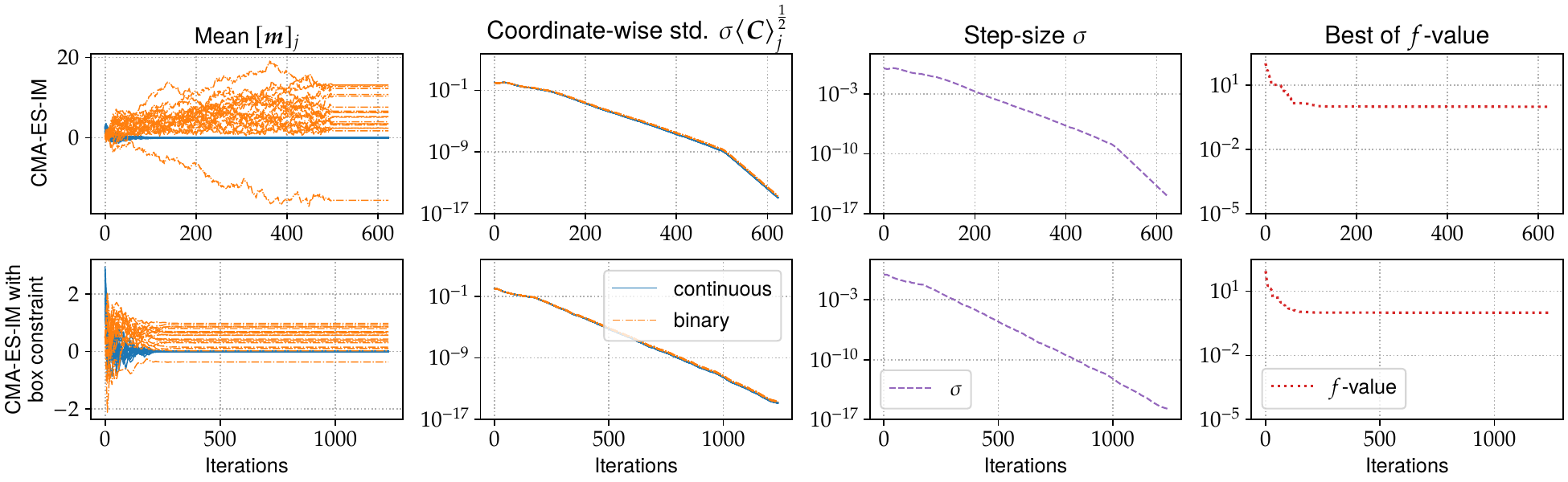}
    \vspace{-2.5mm}
    \caption{Transition of each element of the mean vector and the diagonal elements of the covariance matrix on CMA-ES-IM \cite{hansen_cma-es_2011} with and without the box constraint for a typical single failed trial on 40-dimensional \textsc{SphereOneMax}.}
    \label{fig::transition_m_and_C}
\end{figure*}

\section{Preliminary Experiment: Why Is It Difficult to Optimize Binary Variables?}
\label{sec:preliminary}

It is known that the integer variable handling of CMA-ES \cite{hansen_cma-es_2011} does not work well for binary variables.
However, the reasons for this have not been well explored. We then empirically check why this integer handling fails to optimize binary variables.

We consider the function $\textsc{Encoding}_f(\x_i)$ to binarize the elements of the candidate solution corresponding to the binary variables, and the dimension $N = N_{\mathrm{co}} + N_{\mathrm{bi}}$, where $N_{\mathrm{bi}}$ is the number of binary variables. We define $\textsc{Encoding}_f(\x_i): \mathbb{R}^{N_{\mathrm{co}}} \times \mathbb{R}^{N_{\mathrm{bi}}} \mapsto \mathbb{R}^{N_{\mathrm{co}}} \times \{0, 1 \}^{N_{\mathrm{bi}}}$ as
\begin{align}
    \label{eq::binary_encoding}
    \textsc{Encoding}_f(\x_i) =
    \left\{
        \begin{array}{lll}
            \lbrack \x_i \rbrack_{j} &  (1 \leq j \leq N_{\mathrm{co}}) \\
            \mathds{1} \{ \lbrack \x_i \rbrack_{j} > 0 \} & (N_{\mathrm{co}} + 1 \leq j \leq N)
        \end{array}
    \right.  \enspace .
\end{align}
The partially discretized candidate solution obtained by \eqref{eq::binary_encoding} is denoted by $\bar{\x}_i = \textsc{Encoding}_f(\x_i)$.

Compared to the integer variables, binary variables have a much wider interval, where the same binary variables can be taken after binarization. Therefore, if the variance decreases while the mean vector is so far from the threshold zero at which the binary variable changes, the optimization of the binary variable fails.

\paragraph{Settings}
We use the \textsc{SphereOneMax} function as the objective function, which is a combination of the \textsc{Sphere} function and the \textsc{OneMax} function for the continuous and binary variables, respectively. The \textsc{SphereOneMax} function is defined as
\begin{align}
    \textsc{SphereOneMax}(\bar{\x}_i) = \sum_{j=1}^{N_{\mathrm{co}}} \lbrack \bar{\x}_i\rbrack_j^2 + N_{\mathrm{bi}} - \sum_{k=N_{\mathrm{co}}+1}^N \lbrack \bar{\x}_i \rbrack_k  \enspace, 
\end{align}
where the optimal solution is 0 for continuous variables and 1 for binary variables, respectively, and $\textsc{SphereOneMax}(\bar{\x}^{*}) = 0$. We check the behavior using the CMA-ES with integer variable handling introduced in Section \ref{sec::mi-cma-es} . Additionally, we use this CMA-ES variant with a box constraint $\lbrack \x_i \rbrack_{j} \in \lbrack -1 , 1 \rbrack$ corresponding to the binary variables. When the box constraint is used, the penalty $\| \x_{i}^{\mathrm{feas}} - \x_{i} \|^{2}_{2} / N$ is added to the evaluation value, where $\x_{i}^{\mathrm{feas}}$ is the nearest-neighbor feasible solution to $\x_{i}$. The number of dimensions $N$ is set to 40, and $N_{\mathrm{co}} = N_{\mathrm{bi}} = N/2 = 20$.

The initial mean vector $\m[0]$ is set to uniform random values in the range $\lbrack 1, 3 \rbrack$ for continuous variables and 0 for the binary variables, respectively. The covariance matrix and step-size are initialized with $\C[0] = \boldsymbol{I}$ and $\sig[0] = 1$, respectively. The optimization is successful when the best-evaluated value is less than $10^{-10}$, and the optimization is stopped when the minimum eigenvalue of $\sigma^{2} \boldsymbol{C}$ is less than $10^{-30}$.

\paragraph{Result and Discussion}
For the coordinate-wise mean $\lbrack \boldsymbol{m} \rbrack_{j}$, the coordinate-wise standard deviation $\sigma \langle C \rangle_{j}^{\frac{1}{2}}$, step-size $\sigma$, and best-evaluated value, the upper and lower sides of the Figure \ref{fig::transition_m_and_C} show the transitions of a single typical run of the optimization failure for the CMA-ES with the integer mutation and modification of the step-size adaptation (denoted by CMA-ES-IM) and the CMA-ES-IM with the box constraint. 
The CMA-ES-IM decreases the coordinate-wise standard deviations for binary variables with the step-size. In contrast, coordinate-wise mean is far from the threshold value of zero.
In this case, the integer mutation provided in Step~1 to Step~5 in Section~\ref{sec::mi-cma-es} is not effective to improve the evaluation value because in the dimension corresponding to the binary variable, $\boldsymbol{S}^\mathrm{int} \boldsymbol{r}^\mathrm{int}_i $~($i = 1, \ldots, \lambda_\mathrm{int}^{(t)}$) are smaller than the distances between $\m[t] + \sig[t]\boldsymbol{y}_i$ and the threshold value of zero. Moreover, when $\boldsymbol{S}^\mathrm{int} \boldsymbol{r}^\mathrm{int}_\lambda$ calculated in Step~6 also becomes small, the mutation no longer affects the candidate solutions at all~(after 500 iterations in Figure~\ref{fig::transition_m_and_C}).

On the other hand, CMA-ES-IM with box constraint can prevent the coordinate-wise mean from being far from zero and avoid fixation of candidate solutions by the mutation.
However, even if the mutation works, a high penalty value of box constraint results in a poor evaluation value. In this case, the mutated samples cannot be reflected in the mean update, which uses only the superior $\mu$ samples.
Then, the stagnation problem in the negative domain still remains.

These results suggest that we need a new way to handle integer variables that takes binary variables into account instead of the integer mutation. In Section \ref{sec:proposed}, we propose an integer handling method that preserves the generation probability of a different integer variable by introducing a correction that brings the coordinate-wise mean closer to a threshold value as the coordinate-wise standard deviation decreases.
    
\section{Proposed Method}
\label{sec:proposed}
In this section, we propose a simple modification of the CMA-ES in the MI-BBO. The basic idea is to introduce a lower bound on the marginal probability referred to as the \textit{margin}, so that the sample is not fixed to a single integer variable. The margin is a common technique in 
the estimation of distribution algorithms~(EDAs) for binary domains to address the problem of bits being fixed to 0 or 1. In fact, the population-based incremental learning~(PBIL)~\cite{PBIL:1994}, a binary variable optimization method based on Bernoulli distribution, restricts the updated marginals to the range $[1/N, 1-1/N]$. This prevents the optimization from stagnating with the distribution converging to an undesirable direction before finding the optimum.

To introduce this margin correction to the CMA-ES, we define a diagonal matrix $\boldsymbol{A}$ whose initial value is given by the identity matrix and redefine the MGD that generates the samples as $\mathcal{N}(\boldsymbol{m}, \sigma^2 \boldsymbol{A} \boldsymbol{C} \boldsymbol{A}^\top)$. The margin correction is achieved by correcting $\boldsymbol{A}$ and $\boldsymbol{m}$ so that the probability of the integer variables being generated outside the dominant values is maintained above a certain value $\alpha$. Because the sample generated from $\mathcal{N}(\boldsymbol{m}, \sigma^2 \boldsymbol{A} \boldsymbol{C} \boldsymbol{A}^\top)$ is equivalent to applying the affine transformation of $\boldsymbol{A}$ to the sample generated from $\mathcal{N}(\boldsymbol{m}, \sigma^2 \boldsymbol{C})$, we can separate the adaptation of the covariance and the update of $\boldsymbol{A}$. Consequently, the proposed modification can be represented as the affine transformation of the samples used to evaluate the objective function, without making any changes to the updates in CMA-ES. 
It should be noted that although the mean vector can also be corrected by the affine transformation, we directly correct it to avoid the divergence of $\boldsymbol{m}$.

In this section, we first redefine $\textsc{Encoding}_f$ to facilitate the introduction of the margin. 
Next, we show the process of the CMA-ES with the proposed modification.
Finally, we explain the margin correction, namely, the updates of $\boldsymbol{A}$ and $\boldsymbol{m}$, separately for the cases of binary and integer variables.

The detailed algorithm of the proposed CMA-ES with Margin appears in the supplementary material, and the code is available at \url{https://github.com/EvoConJP/CMA-ES_with_Margin}.

\subsection{Definition of $\textsc{Encoding}_f$ and Threshold $\ell$}
Let $z_{j,k}$ be the $k$-th smallest value among the discrete values in the $j$-th dimension, where $N_{\mathrm{co}} + 1 \leq j \leq N$ and $1 \leq k \leq K_j$.
It should be noted that $K_j$ is the number of candidate integers for the $j$-th variable $\boldsymbol{z}_j$.
Under this definition, the binary variable can also be represented as, e.g. $z_{j,1}=0$, $z_{j,2}=1$. Moreover, we introduce a threshold $\thd$ for encoding continuous variables into discrete variables. Let $\ell_{j, k|k+1}$ be the threshold of two discrete variables $z_{j,k}$ and $z_{j,k+1}$; it is given by the midpoint of $z_{j,k}$ and $z_{j,k+1}$, namely, $\ell_{j, k|k+1} := (z_{j,k} + z_{j,k+1})/2$. We then redefine $\textsc{Encoding}_f$ when $N_{\mathrm{co}} + 1 \leq j \leq N$ as follows:
\begin{align*}
    \textsc{Encoding}_f([\x_i]_j) =
    \left\{
        \begin{array}{lll}
            z_{j,1} & \text{if} \enspace [\x_i]_j \leq \ell_{j, 1|2} \\
            z_{j,k} & \text{if} \enspace \ell_{j, k-1|k} < [\x_i]_j \leq \ell_{j, k|k+1} \\
            z_{j,K_j} & \text{if} \enspace \ell_{j, K_j-1|K_j} < [\x_i]_j
        \end{array}
    \right.
\end{align*}
Moreover, if $1 \leq j \leq N_{\mathrm{co}}$, $[\x_i]_j$ is isometrically mapped as \linebreak $\textsc{Encoding}_f([\x_i]_j) = [\x_i]_j$.
Then, the discretized candidate solution is denoted by $\bar{\x}_i = \textsc{Encoding}_f(\x_i)$. The set of discrete variables $\boldsymbol{z}_j$ is not limited to consecutive integers such as $\{ 0,1,2 \}$, but can also handle general discrete variables such as $\{ 1,2,4 \}$ and $\{ 0.01,0.1,1 \}$.

\subsection{CMA-ES with the Proposed Modification}
Given $\A[0]$ as an identity matrix $\boldsymbol{I}$, the update of the proposed method, termed \textit{CMA-ES with margin}, at the iteration $t$ is given in the following steps.
\begin{enumerate}[\textrm{Step }1\textrm{. }]
    \item The $\lambda$ candidate solutions $\x_i$~($i = 1,2,\ldots, \lambda$) are sampled from $\mathcal{N} (\m[t], (\sig[t])^2 \C[t])$ as $\x_i = \m[t] + \sig[t] \y_i$, where $\y_i \sim \mathcal{N} (\boldsymbol{0}, \C[t])$ for $i = 1,2,\ldots, \lambda$.
    \item The affine transformed solutions $\boldsymbol{v}_i$~($i = 1,2,\ldots, \lambda$) are calculated as $\boldsymbol{v}_i = \m[t] + \sig[t] \A[t] \y_i$ for $i = 1,2,\ldots, \lambda$.
    \item The discretized $\boldsymbol{v}_i$, i.e., $\bar{\boldsymbol{v}}_i$~($i = 1,2,\ldots, \lambda$) are evaluated by $f$ and sort $\{\x_{1:\lambda}, \x_{2:\lambda}, \dots ,\x_{\lambda:\lambda}\}$ and $\{\y_{1:\lambda}, \y_{2:\lambda}, \dots ,\y_{\lambda:\lambda}\}$ so that the indices correspond to $f(\bar{\boldsymbol{v}}_{1:\lambda}) \leq f(\bar{\boldsymbol{v}}_{2:\lambda}) \leq \cdots \leq f(\bar{\boldsymbol{v}}_{\lambda:\lambda})$.
    \item Based on \eqref{eq:cma-es-m} to \eqref{eq:cma-es-sig}, update $\m[t]$, $\C[t]$, and $\sig[t]$ using $\x$ and $\y$.
    \item Modify $\m[t+1]$ and update $\A[t]$ based on Section~\ref{ssec:margin_bin} and Section~\ref{ssec:margin_int}. \label{step:modify}
\end{enumerate}
It should be noted that the algorithm based on the above is consistent with the original CMA-ES if no corrections are made in Step~\ref{step:modify}. In other words, the smaller the margin parameter $\alpha$, described in Section~\ref{ssec:margin_bin} and Section~\ref{ssec:margin_int}, and the more insignificant the modification, the closer the above algorithm is to the original CMA-ES. Moreover, the update of the variance-covariance has not been modified, which facilitates the smooth consideration of the introduction of the CMA-ES properties, e.g., step-size adaptation methods other than CSA.

\begin{figure}[t]
    \begin{center}
      \includegraphics[width=0.80\linewidth]{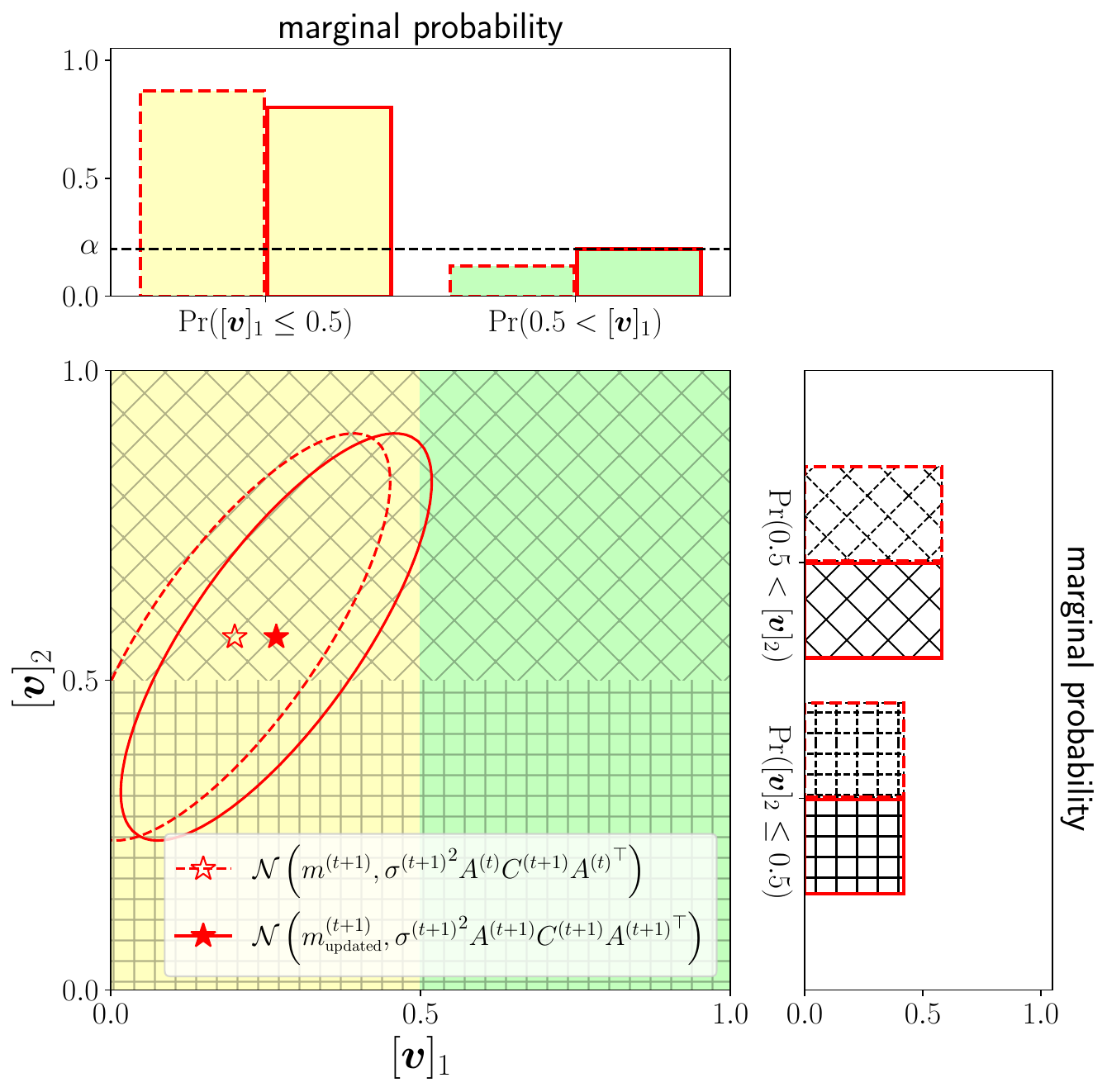}
    \end{center}
    \vspace{-3.0mm}
    \caption{Example of MGD followed by $\boldsymbol{v}$ and its marginal probability. The dashed red ellipse corresponds to the MGD before the correction, whereas the solid one corresponds to the MGD after the correction for the binary variable.}
    \label{fig:bin_correction}
\end{figure}
\subsection{Margin for Binary Variables} \label{ssec:margin_bin}
Considering the probability that a binarized variable $[\bar{\boldsymbol{v}}]_j$ is $0$ and the probability that it is $1$, the following conditions should be satisfied after the modification by the margin:
\begin{multline}
    \min \left\{ \Pr([\bar{\boldsymbol{v}}]_j = 0), \Pr([\bar{\boldsymbol{v}}]_j = 1) \right\} \geq \alpha \\
    \Leftrightarrow \min \left\{ \Pr([\boldsymbol{v}]_j < 0.5), \Pr([\boldsymbol{v}]_j \geq 0.5) \right\} \geq \alpha
\end{multline}
It should be noted that the binarize threshold $\ell_{j,1|2}$ is equal to $0.5$. Here, Figure~\ref{fig:bin_correction} shows an example of the updated MGD followed by $\boldsymbol{v}$ and the marginal probabilities. The MGD before the margin correction~(dashed red ellipse) shows that its marginal probability $\Pr([\boldsymbol{v}]_1 \geq 0.5)$ is smaller than the margin parameter $\alpha$. In this case, by modifying the element of the mean vector $[\m[t+1]]_1$, we can correct the marginal probability $\Pr([\boldsymbol{v}]_1 \geq 0.5)$ to $\alpha$ without affecting the other dimensions. To calculate the amount of the correction for this mean vector, we consider the confidence interval of the probability $1-2\alpha$ in the marginal distribution of the $j$-th dimension. The confidence interval of the $j$-th dimension is represented by
\begin{align*}
    \left[ [\m[t+1]]_j - \textrm{CI}_j^{(t+1)} (1-2\alpha), [\m[t+1]]_j + \textrm{CI}_j^{(t+1)} (1-2\alpha) \right] \enspace.
\end{align*}
It should be noted that $\textrm{CI}_j^{(t+1)} (1-2\alpha)$ is defined as
\begin{align*}
    \textrm{CI}_j^{(t+1)} (1-2\alpha) := \sqrt{\chi^2_{\textrm{ppf}}(1-2\alpha) {\sig[t+1]}^2 \left\langle \A[t] \C[t+1] {\A[t]}^\top\right\rangle_j } \enspace,
\end{align*}
where $\chi^2_\textrm{ppf}(\cdot)$ is the function that, given the lower cumulative probability, returns the percentage point in the chi-squared distribution with $1$ degree of freedom. If the threshold $\ell_{j,1|2} = 0.5$ is outside this confidence interval, the marginal probability to be corrected is less than $\alpha$. Given the encoding threshold closest to $[\m[t+1]]_j$ as $\ell\bigl([\m[t+1]]_j\bigr)$, which is equal to $0.5$ in the binary case, the modification for the $j$-th element of the mean vector can be denoted as
\begin{align}
    &[\m[t+1]]_j \leftarrow \ell\left([\m[t+1]]_j\right) + \sign\left( [\m[t+1]]_j - \ell\left([\m[t+1]]_j\right) \right) \notag \\
    &\quad \cdot \min \left\{ \left| [\m[t+1]]_j - \ell\left([\m[t+1]]_j\right) \right|, \textrm{CI}_j^{(t+1)} (1-2\alpha) \right\} \enspace. \label{eq:bin_correct_m}
\end{align}
Additionally, no changes are made to $\langle \A[t] \rangle_j$, namely,
\begin{align}
   \langle \A[t+1] \rangle_j \leftarrow \langle \A[t] \rangle_j \enspace. \label{eq:bin_correct_A}
\end{align}
As shown in the solid red line in Figure~\ref{fig:bin_correction}, the marginal probability after this modification is lower-bounded by $\alpha$.

\subsection{Margin for Integer Variables} \label{ssec:margin_int}
First, we consider the cases where the $j$-th element of the mean vector satisfies $[\m[t+1]]_j \leq \ell_{j, 1|2}$ or $\ell_{j, K_j-1|K_j} < [\m[t+1]]_j$. In these cases, the integer variable $[\boldsymbol{v}]_j$ may be fixed to $z_{j,1}$ or $z_{j,K_j}$, respectively. Thus, we correct the marginal probability of generating one inner integer variable, i.e., $z_{j,2}$ for $z_{j,1}$ or $z_{j,K_j-1}$ for $z_{j,K_j}$, to maintain it above $\alpha$, respectively. This correction is achieved by updating $[\m[t+1]]_j$ and $\A[t]$ based on \eqref{eq:bin_correct_m} and \eqref{eq:bin_correct_A}.

\begin{figure}[t]
    \begin{center}
      \includegraphics[width=0.80\linewidth]{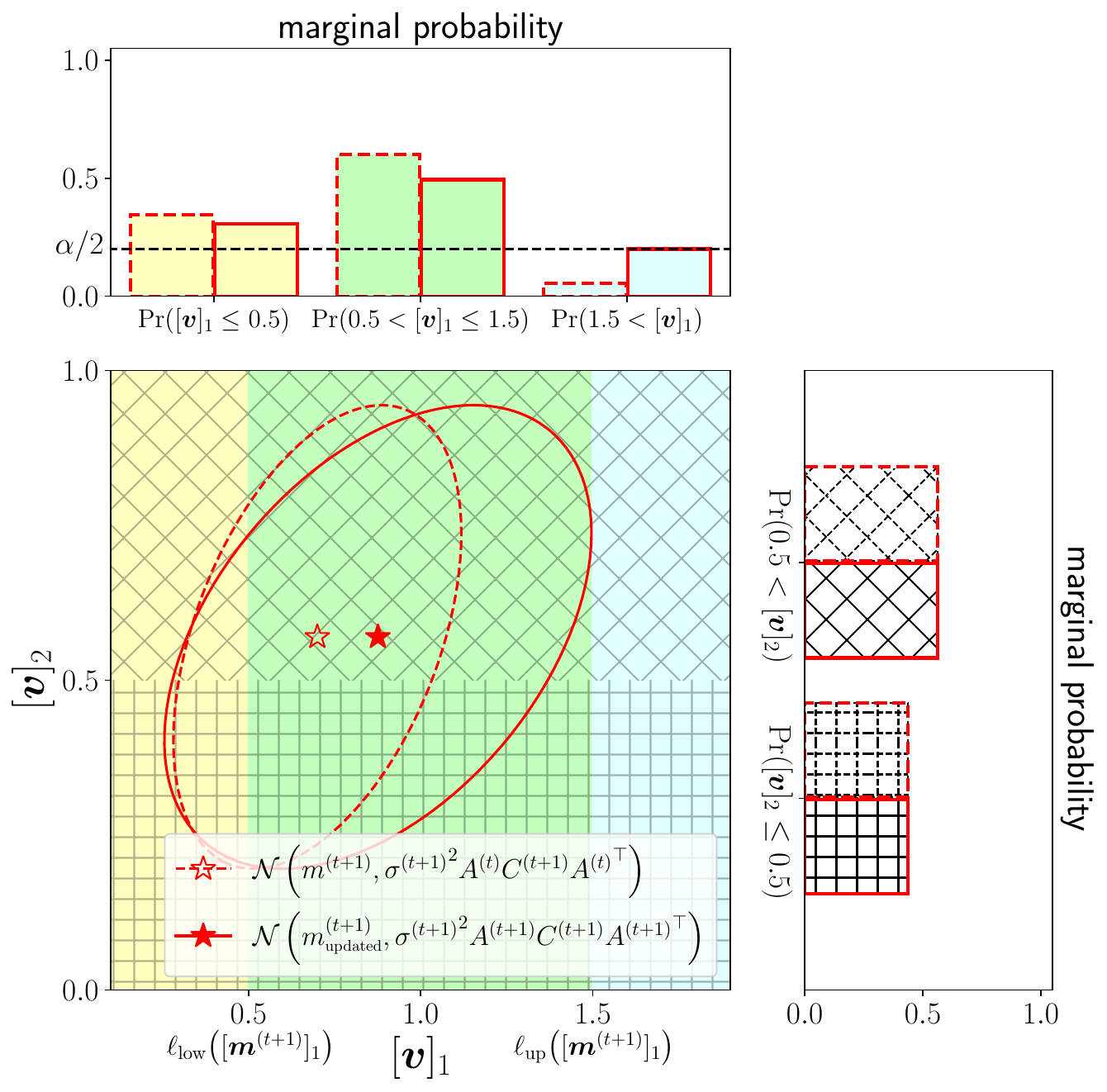}
    \end{center}
    \vspace{-3.0mm}
    \caption{Example of MGD followed by $\boldsymbol{v}$ and its marginal probability. The dashed red ellipse corresponds to the MGD before the correction, whereas the solid one corresponds to the MGD after the correction for the integer variable.}
    \label{fig:int_correction}
\end{figure}

Next, we consider the case of other integer variables. Figure~\ref{fig:int_correction} shows an example of the updated MGD followed by $\boldsymbol{v}$ and the marginal probability. In this example, $[\boldsymbol{v}]_1$ is expected to be fixed in the interval $(0.5,1.5]$ when $\Pr([\boldsymbol{v}]_1 \leq 0.5)$ and $\Pr(1.5 < [\boldsymbol{v}]_1)$ become small. Thus, the correction strategy is to lower-bound the probability of $[\boldsymbol{v}]_j$ being generated outside the plateau where $[\boldsymbol{v}]_j$ is expected to be fixed, such as $\Pr([\boldsymbol{v}]_1 \leq 0.5)$ and $\Pr(1.5 < [\boldsymbol{v}]_1)$ in Figure~\ref{fig:int_correction}. In this case, the value of the margin is set to $\alpha/2$.
For simplicity, we denote $\ell_\textrm{low}\bigl([\m[t+1]]_j\bigr)$ and $\ell_\textrm{up}\bigl([\m[t+1]]_j\bigr)$ as
\begin{align}
    \ell_\textrm{low}\left([\m[t+1]]_j\right) &:= \max \left\{ l \in \thd_j : l < [\m[t+1]]_j \right\} \enspace, \\
    \ell_\textrm{up}\left([\m[t+1]]_j\right) &:= \min \left\{ l \in \thd_j : [\m[t+1]]_j \leq l \right\} \enspace.
\end{align}
The first step of the modification is to calculate $p_\textrm{low}$, $p_\textrm{up}$, and $p_\textrm{mid}$ as follows.\shin{Please check this sentence; $p_\textrm{low}$ and $p_\textrm{up}$ are duplicated.}\hamano{checked and modified.}
\begin{align}
    p_\textrm{low} &\leftarrow \Pr\left([\boldsymbol{v}]_j \leq \ell_\textrm{low}\left([\m[t+1]]_j\right)\right) \\
    p_\textrm{up} &\leftarrow \Pr\left(\ell_\textrm{up}\left([\m[t+1]]_j \right) < [\boldsymbol{v}]_j \right) \\
    p_\textrm{mid} &\leftarrow 1 - p_\textrm{low} - p_\textrm{up}
\end{align}
Next, we restrict $p_\textrm{low}$, $p_\textrm{up}$, and $p_\textrm{mid}$ as follows. 
\begin{align}
    p'_\textrm{low} &\leftarrow \max \{ \alpha/2 , p_\textrm{low} \} \label{eq:lower_low} \\
    p'_\textrm{up} &\leftarrow \max \{ \alpha/2 , p_\textrm{up} \} \label{eq:lower_up} \\
    p''_\textrm{low} &\leftarrow p'_\textrm{low} + \frac{1 - p'_\textrm{low} - p'_\textrm{up} - p_\textrm{mid}}{p'_\textrm{low} + p'_\textrm{up} + p_\textrm{mid} - 3 \cdot \alpha/2}(p'_\textrm{low} - \alpha/2) \label{eq:ensure_low} \\
    p''_\textrm{up} &\leftarrow p'_\textrm{up} + \frac{1 - p'_\textrm{low} - p'_\textrm{up} - p_\textrm{mid}}{p'_\textrm{low} + p'_\textrm{up} + p_\textrm{mid} - 3 \cdot \alpha/2}(p'_\textrm{up} - \alpha/2) \label{eq:ensure_up}
\end{align}
The equations \eqref{eq:ensure_low} and \eqref{eq:ensure_up} ensure $p''_\textrm{low} + p''_\textrm{up} + p'_\textrm{mid} = 1$, while keeping $p''_\textrm{low} \geq \alpha/2$ and $p''_\textrm{up} \geq \alpha/2$\new{, where $p'_\textrm{mid}=1-p''_\textrm{low} - p''_\textrm{up}$}. This handling method is also adopted in~\cite[Appendix~D]{ASNG:2019}.
We update $\m[t+1]$ and $\A[t]$ so that the corrected marginal probabilities $\Pr\bigl([\boldsymbol{v}]_j \leq \ell_\textrm{low}\bigl([\m[t+1]]_j\bigr)\bigr)$ and $\Pr\bigl(\ell_\textrm{up}\bigl([\m[t+1]]_j \bigr) < [\boldsymbol{v}]_j \bigr)$ are $p''_\textrm{low}$ and $p''_\textrm{up}$, respectively. The conditions to be satisfied are as follows.
\begin{align}
    \begin{cases}
        [\m[t+1]]_j - \ell_\textrm{low}\left( [\m[t+1]]_j \right) = \textrm{CI}_j^{(t+1)} (1 - 2p''_\textrm{low}) \\
        \ell_\textrm{up}\left( [\m[t+1]]_j \right) - [\m[t+1]]_j = \textrm{CI}_j^{(t+1)} (1 - 2p''_\textrm{up})
    \end{cases} \label{eq:simult}
\end{align}
Finally, the solutions of the simultaneous linear equations for \linebreak $[\m[t+1]]_j$ and $\langle \A[t] \rangle_j$ are applied to the updated $[\m[t+1]]_j$ and $\langle \A[t+1] \rangle_j$. Correcting $\m[t+1]$ and $\A[t]$ in this way bounds both $\Pr\bigl([\boldsymbol{v}]_j \leq  \ell_\textrm{low} \bigr. \bigl. \bigl([\m[t+1]]_j\bigr)\bigr)$ and $\Pr\bigl(\ell_\textrm{up}\bigl([\m[t+1]]_j \bigr) < [\boldsymbol{v}]_j \bigr)$ above $\alpha/2$, as indicated by the solid line in Figure~\ref{fig:int_correction}. Moreover, we note that there are cases where $p_\textrm{mid}$, $\Pr(0.5 < [\m[t+1]]_1 \leq 1.5)$ in Figure~\ref{fig:int_correction}, is less than $\alpha/2$ even with the margin. In that case, the variance is sufficiently large that no fixation of the discrete variable occurs in the corresponding dimension.

\begin{figure*}[tbph]
    \centering
    \includegraphics[width=0.80\linewidth]{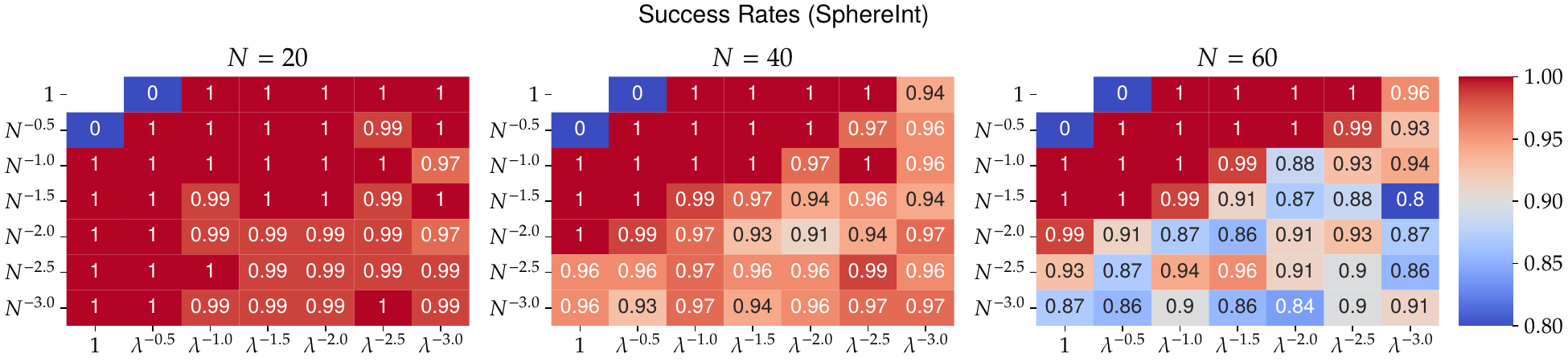}
    \includegraphics[width=0.80\linewidth]{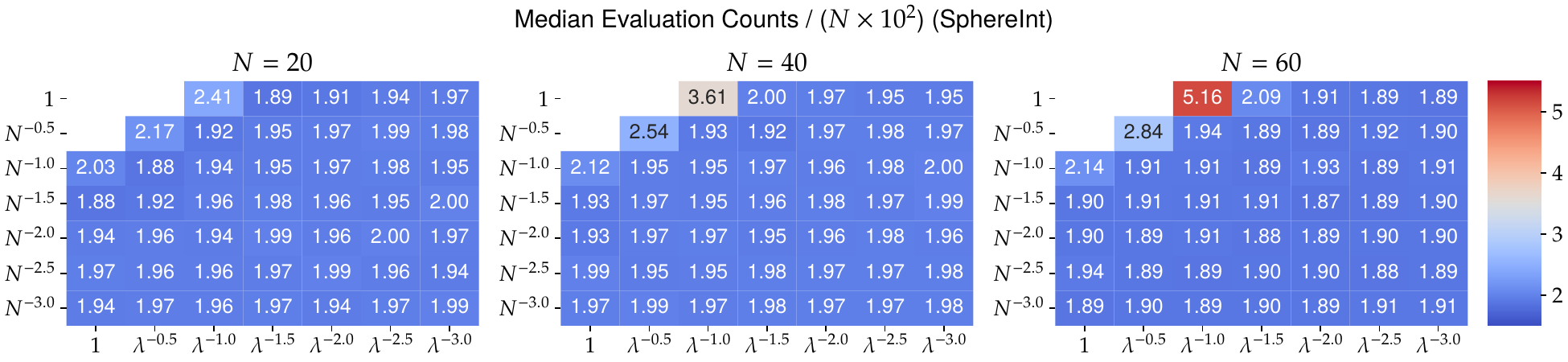}
    \vspace{-2.5mm}
    \caption{Heatmap of the success rate (top) and the median evaluation counts for successful cases (bottom) in the $N$-dimensional \textsc{SphereInt} function when the hyperparameter $\alpha=N^{-m}\lambda^{-n}$ of the proposed method is changed.}
    \label{fig:grid_seaech_success}
\end{figure*}

\section{Experiment and Result}
\label{sec:exp_and_res}
We apply the proposed method to the MI-BBO optimization problem for several benchmark functions to validate its robustness and efficiency. In Section \ref{sec:alpha_search}, we check the performance changes of the proposed method according to the hyperparameter $\alpha$. In Section \ref{sec:benchmark}, we check the difference in the search success rate and the number of evaluations between the proposed method and CMA-ES-IM for several artificial MI-BBO benchmark functions. The definitions of the benchmark functions used in this section\del{Sections \ref{sec:alpha_search} and \ref{sec:benchmark}}{} are listed as below:
\begin{itemize}
    \item  $\textsc{SphereOneMax}(\bar{x}) = \sum_{j=1}^{N_{\mathrm{co}}} \lbrack \bar{\x} \rbrack_j^2 + N_{\mathrm{bi}} - \sum_{k=N_{\mathrm{co}}+1}^N \lbrack \bar{\x} \rbrack_k$
    \item $\textsc{SphereLeadingOnes}(\bar{x}) = \sum_{j=1}^{N_{\mathrm{co}}}\lbrack \bar{\x} \rbrack_j^2 \\ + N_{\mathrm{bi}} - \sum_{k=N_{\mathrm{co}}+1}^N \prod_{l = N_{\mathrm{co}}+1}^k \lbrack \bar{\x} \rbrack_l$
    \item $\textsc{EllipsoidOneMax}(\bar{x}) = \sum_{j=1}^{N_{\mathrm{co}}}\left(1000^{\frac{j-1}{N_{\mathrm{co}}-1}}\lbrack \bar{\x} \rbrack_j \right)^2 \\+ N_{\mathrm{bi}} - \sum_{k=N_{\mathrm{co}}+1}^N \lbrack \bar{\x} \rbrack_k$
    \item $\textsc{EllipsoidLeadingOnes}(\bar{x}) = \sum_{j=1}^{N_{\mathrm{co}}}\left(1000^{\frac{j-1}{N_{\mathrm{co}}-1}}\lbrack \bar{\x} \rbrack_j \right)^2 \\+ N_{\mathrm{bi}} - \sum_{k=N_{\mathrm{co}}+1}^N \prod_{l = N_{\mathrm{co}}+1}^k \lbrack \bar{\x} \rbrack_l$
    \item $\textsc{SphereInt}(\bar{x}) = \sum_{j=1}^{N}\lbrack \bar{\x} \rbrack_j^2$
    \item $\textsc{EllipsoidInt}(\bar{x}) = \sum_{j=1}^{N}\left(1000^{\frac{j-1}{N-1}}\lbrack \bar{\x} \rbrack_j \right)^2$
\end{itemize}
In all functions, the first $N_{\mathrm{co}}$ variables are continuous, whereas the last $N - N_{\mathrm{co}}$ variables are binary or integer.
In all the experiments, we adopted the default parameters of CMA-ES listed in Table \ref{tb:cma_params}.

\begin{table*}[tbph]
  \caption{Comparison of evaluation counts and success rate for the benchmark functions. The evaluation counts are reported as the median value of successful trials. The bold fonts represent the best median evaluation counts and the best success rates among the methods. The inside of the parentheses represents the interquartile range (IQR).}
  \vspace{-2.5mm}
  \label{table:result_benchmark_funcion}
  \centering
  \begin{tabular}{c|c|cc|cc|cc}
    \toprule
    \multirow{3}{*}{Function} & \multirow{3}{*}{$N$} & \multicolumn{2}{c}{CMA-ES-IM} & \multicolumn{2}{|c|}{CMA-ES-IM \& Box-constraint} & \multicolumn{2}{c}{CMA-ES w. Margin (Proposed)} \\
    & & Evaluation & Success & Evaluation & Success & Evaluation  & Success \\
    & & Counts & Rates & Counts & Rates & Counts & Rates \\
    \midrule
    \multirow{3}{*}{\textsc{SphereOneMax}}
        & 20 & \textbf{2964} (174) & 83/100 & 4752 (100) & \textbf{100/100} & 3876 (435) & \textbf{100/100} \\
        & 40 & \textbf{5745} (558) & 62/100 & 7575 (543) & 64/100 & 7995 (514) & \textbf{100/100} \\
        & 60 & \textbf{8112} (340) & 46/100 & 12240 (336) & 21/100 & 12408 (1012) & \textbf{100/100} \\
    \midrule
    \multirow{3}{*}{\textsc{SphereLeadingOnes}} 
        & 20 & \textbf{2904} (192) & 77/100 & 5124 (296) & \textbf{100/100} & 4158 (339) & \textbf{100/100} \\
        & 40 & \textbf{5647} (296) & 24/100 & 8280 (112) & 10/100 & 8505 (724) & \textbf{100/100} \\
        & 60 & \textbf{8816} (12) & 8/100 & 12624 (96) & 2/100 & 13424 (1008) & \textbf{100/100} \\
    \midrule
    \multirow{3}{*}{\textsc{EllipsoidOneMax}}
        & 20 & \textbf{9918} (396) & 20/100 & 26700 (7020) & 96/100 & 11172 (666) & \textbf{100/100} \\
        & 40 & \textbf{35325} (0) & 1/100 & 79912 (28661) & 14/100 & 40590 (1789) & \textbf{100/100} \\
        & 60 & \textbf{78560} (0) & 5/100 & 283440 (0) & 1/100 & 88064 (3536) & \textbf{100/100} \\
    \midrule
    \multirow{3}{*}{\textsc{EllipsoidLeadingOnes}} 
        & 20 & \textbf{10104} (441) & 14/100 & 24880 (7305) & 92/100 & 11454 (876) & \textbf{100/100} \\
        & 40 & - & 0/100 & 85807 (13335) & 4/100 & \textbf{41048} (1744) & \textbf{100/100} \\
        & 60 & - & 0/100 & - & 0/100 & \textbf{91496} (3488) & \textbf{100/100} \\
    \midrule
    \multirow{3}{*}{\textsc{SphereInt}}
        & 20 & 5130 (477) & 86/100 & 5280 (744) & 89/100 & \textbf{3840} (306) & \textbf{100/100} \\
        & 40 & 7950 (697) & 71/100 & 8070 (555) & 85/100 & \textbf{7838} (458) & \textbf{100/100} \\
        & 60 & 13184 (816) & 41/100 & 12992 (544) & 34/100 & \textbf{11512} (544) & \textbf{100/100} \\
    \midrule
    \multirow{3}{*}{\textsc{EllipsoidInt}} 
        & 20 & 19128 (3192) & 76/100 & 19476 (5028)  & 73/100 & \textbf{8418} (837) & \textbf{100/100} \\
        & 40 & 43935 (4095) & 73/100 & 43440 (4762) & 83/100 & \textbf{22815} (1733) & \textbf{100/100} \\
        & 60 & 89200 (6152) & 54/100 & 86848 (6136) & 54/100 & \textbf{42000} (3320) & \textbf{100/100} \\
    \bottomrule
  \end{tabular}
\end{table*}

\subsection{Hyperparameter Sensitivity for $\alpha$}
\label{sec:alpha_search}
We use the \textsc{SphereInt} function of the objective function and adopt the same initialization for the distribution parameters and termination condition as in Section \ref{sec:preliminary}. The number of dimensions $N$ is set to 20, 40, or 60, and the numbers of the continuous and integer variables are $N_{\mathrm{co}} = N_{\mathrm{int}} = N/2$. The integer variables are assumed to take values in the range $\lbrack -10, 10 \rbrack$. 
We argue that it is reasonable that the hyperparameter $\alpha$, which determines the margin in the proposed method, should depend on the number of dimensions $N$ and the sample size $\lambda$. In this experiment, we evaluate a total of 48 settings except for $\alpha = 1$ which we set as $\alpha = N^{-m} \lambda^{-n} (m, n \in \lbrack 0, 0.5, 1, 1.5, 2, 2.5, 3 \rbrack)$. In each setting, 100 trials are performed independently using different seed values. 

\paragraph{Results and Discussion}
Figure~\ref{fig:grid_seaech_success} shows the success rate and the median evaluation count for successful cases in each setting. For the success rate, when $\alpha$ is set to a large value as $N^{-0.5}$ or $\lambda^{-0.5}$, all trials fail. Additionally, when $\alpha$ is set as smaller than $(N\lambda)^{-1}$, the success rate tends to decrease as the number of dimensions $N$ increases. If $\alpha$ is too large, the probability of the integer changing is also too large and the optimization is unstable;
however, if $\alpha$ is too small, it is difficult to get out of the stagnation because the conditions under which the mean $[\boldsymbol{m}]_{j}$ and affine matrix $\boldsymbol{A}$ corrections are applied become more stringent. 
For the median evaluation count, there is no significant difference for any dimension except for $\alpha = N^{-1}, (N \lambda)^{-0.5}, \lambda^{-1}$. 
Therefore, for robustness and efficiency reasons, we use $\alpha = (N\lambda)^{-1}$ as a default parameter in the subsequent experiments in this study.

\subsection{Comparison of Optimization Performance}
\label{sec:benchmark}
We compare the optimization performance of the benchmark functions listed in Section \ref{sec:exp_and_res} for the proposed method, CMA-ES-IM, and CMA-ES-IM with box constraints. As in Section \ref{sec:alpha_search}, the number of dimensions $N$ is set to 20, 40, and 60. The number of continuous and integer variables are $N_{\mathrm{co}} = N_{\mathrm{bi}} = N_{\mathrm{int}} = N/2$, respectively. For CMA-ES-IM with the box constraint, \textsc{SphereOneMax}, \textsc{SphereLeadingOnes}, \textsc{EllipsoidOneMax}, and \textsc{ElipsoidLeadingOnes} functions are given the constraint $\lbrack \x \rbrack_{j} \in \lbrack -1, 1 \rbrack$ corresponding to the binary variables, and other functions are given the constraint $\lbrack \x \rbrack_{j} \in \lbrack -10, 10 \rbrack$ corresponding to the integer variables. The calculating method of the penalty for violating the constraints is the same as in Section \ref{sec:preliminary}. The optimization is successful when the best-evaluated value is less than $10^{-10}$, whereas the optimization is stopped when the minimum eigenvalue of $\sigma^{2} \boldsymbol{C}$ is less than $10^{-30}$ or the condition number of $\boldsymbol{C}$ exceeds $10^{14}$.

\paragraph{Results and Discussion}
Table \ref{table:result_benchmark_funcion} summarizes the median evaluation counts and success rates in each setting. Comparing the proposed method with the CMA-ES-IM with and without the box constraint for the \textsc{SphereOneMax} and \textsc{SphereLeadingOnes} functions, the CMA-ES-IM reaches the optimal solution in fewer evaluation counts. However, the success rate of CMA-ES-IM with and without the box constraint decreases as the number of dimensions increases. However, the success rate of the proposed method remains 100\% regardless of the increase in the number of dimensions. 
For the \textsc{EllipsoidOneMax} and \textsc{EllipsoidLeadingOnes} functions, the CMA-ES-IM without the box constraint fails on most of the trials in all dimensions, and the CMA-ES-IM with box constraint has a relatively high success rate in $N = 20$ but deteriorates rapidly in $N = 40$ or more dimensions.
In contrast, the proposed method maintains a 100\% success rate and reaches the optimal solution in fewer evaluation counts than the CMA-ES-IM with the box constraint in $N = 20, 40$. For the \textsc{SphereInt} and \textsc{EllipsoidInt} functions, the proposed method successfully optimizes with fewer evaluations in all dimensions than the other methods, maintaining a 100\% success rate. These results show that the proposed method can perform the MI-BBO robustly and efficiently for multiple functions with an increasing number of dimensions. 

\section{Conclusion}
\label{sec:conclusion}
In this work, we first experimentally confirmed that the existing integer handling method, CMA-ES-IM~\cite{hansen_cma-es_2011} with or without the box constraint, does not work effectively for binary variables, and then proposed a new integer variable handling method for CMA-ES. In the proposed method, the mean vector and the diagonal affine transformation matrix for the covariance matrix are corrected so that the marginal probability for an integer variable is lower-bounded at a certain level, which is why the proposed method is called the CMA-ES with \textit{margin}; it considers both the binary and integer variables.

The proposed method has a hyperparameter, $\alpha$, that determines the degree of the lower bound for the marginal probability. We investigated the change in the optimization performance on the \textsc{SphereInt} function with multiple $\alpha$ settings in order to determine the default parameter. With the recommended value of $\alpha$, we experimented the proposed method on several MI-BBO benchmark problems. The experimental results demonstrated that the proposed method is robust even when the number of dimensions increases and can find the optimal solution with fewer evaluations than the existing method, CMA-ES-IM with or without the box constraint.

There are still many challenges left for the MI-BBO; for example, \citet{tusar_mixed-integer_2019} pointed out the difficulty of optimization for non-separable ill-conditioned convex-quadratic functions, such as the rotated Ellipsoid function. In future, we need to address these issues, which have not yet been addressed by the proposed or existing methods, by considering multiple dimension correlations. Additionally, evaluating the proposed method on real-world MI-BBO problems is also an important future direction.


\begin{acks}
The authors thank anonymous reviewers for their helpful comments.
This work was partially supported by JSPS KAKENHI Grant Number JP20H04240.
\end{acks}

\bibliographystyle{ACM-Reference-Format}
\bibliography{reference}

\newpage
\appendix
\input{appendix}

\end{document}

%% file: appendix.tex
\clearpage
\onecolumn
\begin{algorithm}[b]
    \caption{Single update in CMA-ES with Margin for optimization problem $\min_{\x} f(\x)$}
    \label{alg:proposed}
    \begin{algorithmic}[1]
        \STATE \textbf{given } $\m[t] \in \R^N$, $\sig[t] \in \R_{+}$, $\C[t] \in \R^{N \times N}$, $\ps[t] \in \R^N$, $\pc[t] \in \R^N$, and $\A[t] \in \R^{N \times N}$~(diagonal matrix)
        \FOR {$i = 1, \ldots, \lambda$}
            \STATE $\y_i \sim \mathcal{N}(\boldsymbol{0}, \C[t])$
            \STATE $\x_i \leftarrow \m[t] + \sig[t] \y_i$
            \STATE $\boldsymbol{v}_i \leftarrow \m[t] + \sig[t] \A[t] \y_i^\top$
            \STATE $\bar{\boldsymbol{v}}_i \leftarrow \textsc{Encoding}_f(\boldsymbol{v}_i)$
        \ENDFOR
        \STATE Sort $\{\x_{1:\lambda}, \x_{2:\lambda}, \ldots, \x_{\lambda:\lambda}\}$ and $\{\y_{1:\lambda}, \y_{2:\lambda}, \ldots, \y_{\lambda:\lambda}\}$ so that the indices correspond to $f(\bar{\boldsymbol{v}}_{1:\lambda}) \leq f(\bar{\boldsymbol{v}}_{2:\lambda}) \leq \ldots \leq f(\bar{\boldsymbol{v}}_{\lambda:\lambda})$
        \STATE $\m[t+1] \leftarrow \m[t] + c_{m} \sum_{i=1}^{\mu} w_{i} ( \x_{i:\lambda} - \m[t] )$
        \STATE $\ps[t+1] \leftarrow (1-c_\sigma)\ps[t] + \sqrt{c_\sigma(2-c_\sigma)\muw} {\C[t]}^{-\frac12} \sum_{i=1}^\mu \w_i \y_{i:\lambda}$
        \STATE $h_\sigma \leftarrow \mathds{1} {\{\|\ps[t+1]\| < \sqrt{1-(1-c_\sigma)^{2(t+1)}}\left(1.4+\frac{2}{N+1}\right)\E [\|\mathcal{N}(\boldsymbol{0}, \boldsymbol{I}) \|]\}}$
        \STATE $\pc[t+1] \leftarrow (1-c_c) \pc[t] + h_\sigma \sqrt{c_c(2-c_c)\muw} \sum_{i=1}^\mu \w_i \y_{i:\lambda}$
        \STATE $\C[t+1] \leftarrow \left(1 - c_1 - c_\mu \sum_{i=1}^\lambda w_i + (1-h_\sigma)c_1 c_c(2-c_c) \right) \C[t] + \underbrace{c_1 \pc[t+1]{\pc[t+1]}^\top}_{\text{rank-one update}} + \underbrace{c_\mu \sum_{i=1}^\lambda w_i^{\circ} \y_{i:\lambda}\y_{i:\lambda}^\top}_{\text{rank-}\mu\text{ update}}$
        \STATE $\sig[t+1] \leftarrow \sig[t] \exp \left( \frac{c_\sigma}{d_\sigma} \left( \frac{\|\ps[t+1] \|}{\E [\|\mathcal{N}(\boldsymbol{0}, \boldsymbol{I}) \|]} - 1 \right) \right)$
        \STATE \texttt{// Margin for Binary Variables}
        \FOR {$j = N_{\mathrm{co}}+1, \ldots, N_{\mathrm{co}}+N_{\mathrm{bi}}$}
            \STATE $[\m[t+1]]_j \leftarrow \ell\left([\m[t+1]]_j\right) + \sign\left( [\m[t+1]]_j - \ell\left([\m[t+1]]_j\right) \right) \min \left\{ \left| [\m[t+1]]_j - \ell\left([\m[t+1]]_j\right) \right|, \textrm{CI}_j^{(t+1)} (1-2\alpha) \right\}$
            \STATE $\langle \A[t+1] \rangle_j \leftarrow \langle \A[t] \rangle_j$
        \ENDFOR
        \STATE \texttt{// Margin for Integer Variables}
        \FOR {$j = N_{\mathrm{co}}+N_{\mathrm{bi}}+1, \ldots, N$}
            \IF {$[\m[t+1]]_j \leq \ell_{j,1|2}$ or $\ell_{j,K_j-1|K_j} < [\m[t+1]]_j$}
                \STATE $[\m[t+1]]_j \leftarrow \ell\left([\m[t+1]]_j\right) + \sign\left( [\m[t+1]]_j - \ell\left([\m[t+1]]_j\right) \right) \min \left\{ \left| [\m[t+1]]_j - \ell\left([\m[t+1]]_j\right) \right|, \textrm{CI}_j^{(t+1)} (1-2\alpha) \right\}$
                \STATE $\langle \A[t+1] \rangle_j \leftarrow \langle \A[t] \rangle_j$
            \ELSE
                \STATE $[\m[t+1]]_j \leftarrow \frac{\ell_\textrm{low}\left([\m[t+1]]_j\right) \sqrt{\chi^2_{\textrm{ppf}}(1-2p''_\textrm{up})} + \ell_\textrm{up}\left([\m[t+1]]_j\right) \sqrt{\chi^2_{\textrm{ppf}}(1-2p''_\textrm{low})}}{\sqrt{\chi^2_{\textrm{ppf}}(1-2p''_\textrm{low})} + \sqrt{\chi^2_{\textrm{ppf}}(1-2p''_\textrm{up})}}$
                \STATE $\langle \A[t+1] \rangle_j \leftarrow \frac{\ell_\textrm{up}\left([\m[t+1]]_j\right) - \ell_\textrm{low}\left([\m[t+1]]_j\right)}{\sigma^{(t+1)} \sqrt{\langle \C[t+1] \rangle_j} \left( \sqrt{\chi^2_{\textrm{ppf}}(1-2p''_\textrm{low})} + \sqrt{\chi^2_{\textrm{ppf}}(1-2p''_\textrm{up})} \right) }$
            \ENDIF
        \ENDFOR
    \end{algorithmic}
\end{algorithm}

\begin{multicols}{2}
\section{Algorithm Details of the CMA-ES with Margin} \label{apdx}
First, we show the updated $[\m[t+1]]_j$ and $\langle \A[t+1] \rangle_j$ in the margin for the integer variables.
Solving the simultaneous linear equations in \eqref{eq:simult} for $[\m[t+1]]_j$ and $\langle \A[t] \rangle_j$, we obtain $[\m[t+1]]_j$ as
{\small
\begin{align*}
    \frac{\ell_\textrm{low}\left([\m[t+1]]_j\right) \sqrt{\chi^2_{\textrm{ppf}}(1-2p''_\textrm{up})} + \ell_\textrm{up}\left([\m[t+1]]_j\right) \sqrt{\chi^2_{\textrm{ppf}}(1-2p''_\textrm{low})}}{\sqrt{\chi^2_{\textrm{ppf}}(1-2p''_\textrm{low})} + \sqrt{\chi^2_{\textrm{ppf}}(1-2p''_\textrm{up})}}
\end{align*}
}
and $\langle \A[t] \rangle_j$ as 
{\small
\begin{align*}
    \frac{\ell_\textrm{up}\left([\m[t+1]]_j\right) - \ell_\textrm{low}\left([\m[t+1]]_j\right)}{ \sigma^{(t+1)} \sqrt{\langle \C[t+1] \rangle_j} \left( \sqrt{\chi^2_{\textrm{ppf}}(1-2p''_\textrm{low})} + \sqrt{\chi^2_{\textrm{ppf}}(1-2p''_\textrm{up})} \right) } \enspace.
\end{align*}
}These solutions are applied as updated $[\m[t+1]]_j$ and $\langle \A[t+1] \rangle_j$, respectively.

Finally, the single update in the CMA-ES with Margin is shown in Algorithm~\ref{alg:proposed}. Note that here we consider a minimization problem $\min_{\x} f(\x)$, where $[\x]_j~(j = 1, \ldots, N_{\mathrm{co}})$ are continuous variables, $[\x]_j~(j = N_{\mathrm{co}} + 1, \ldots, N_{\mathrm{co}}+N_{\mathrm{bi}})$ are binary variables, and $[\x]_j~(j = N_{\mathrm{co}}+N_{\mathrm{bi}} + 1, \ldots, N)$ are integer variables.
\end{multicols}